\documentclass{article} 
\usepackage[T1]{fontenc}
\usepackage{iclr2027_conference,times}

\usepackage{amsmath,amsfonts,bm}

\def\eqref#1{equation~\ref{#1}}

\def\1{\bm{1}}

\DeclareMathAlphabet{\mathsfit}{\encodingdefault}{\sfdefault}{m}{sl}
\SetMathAlphabet{\mathsfit}{bold}{\encodingdefault}{\sfdefault}{bx}{n}

\DeclareMathOperator*{\argmax}{arg\,max}

\usepackage[table]{xcolor}
\usepackage{hyperref}
\usepackage{url}
\usepackage{graphicx}
\usepackage{booktabs}
\usepackage{multirow}
\usepackage{adjustbox}
\usepackage{amsmath}
\usepackage{amssymb}
\usepackage{algorithm}
\usepackage{algpseudocode}
\usepackage{float}
\usepackage{capt-of}

\definecolor{myblue}{HTML}{4E95D9}
\definecolor{upgreen}{HTML}{2E9E5B}   
\definecolor{downred}{HTML}{D9534F}   

\newcommand{\ours}{VidHarness}

\title{\ours{}: Evolving Agent Harnesses for\\ Cost-Efficient Long Video Understanding}

\author{Susan Liang$^1$, Jianmin Wu$^1$, Daxiang Dong$^1$ \\
$^1$Baidu, Inc.\\
\texttt{\{liangsusan,wujianmin,dongdaxiang\}@baidu.com} \\
}

\iclrfinalcopy 
\begin{document}

\maketitle

\begin{abstract}
Vision-language models (VLMs) can answer questions about hour-long videos, but processing every frame is prohibitively expensive, even though the evidence for a question usually spans only a few seconds. Video agents, i.e., harness programs wrapped around a frozen VLM, address this by observing the video selectively, yet existing harnesses are hand-crafted by experts through slow build-and-test cycles. We propose \ours{}, a framework that automates harness design for cost-efficient long video understanding, in which a harness proposer iteratively evolves harnesses based on execution feedback from an evolution environment. To escape the local optima of greedy refinement, we organize the evolution as Monte Carlo tree search (MCTS), and to reduce the evaluation cost, we integrate uncertainty-aware multi-fidelity validation, which screens new harnesses on a few questions and promotes only the promising ones. Since the best harness varies with the frame budget, we further introduce a mixture-of-harness that routes each question to a harness specialized for its budget. \ours{} sets new state-of-the-art results on LongVideoBench, Video-MME, and Video-Holmes, outperforms the strongest hand-crafted video agent by up to $11.2$ points, and generalizes to the knowledge-intensive benchmarks Video-MMMU and MMVU with fewer than half of the frames of uniform sampling.
\end{abstract}

\section{Introduction}
\label{sec:intro}

Machine perception is shifting from clips to hours. Film and sports analysis must follow narratives across full-length footage~\citep{wu2024longvideobench, wang2025lvbench, fu2025video, cheng2025video}, personal assistants must recall events from day-long egocentric recordings~\citep{tian2026ego}, and surveillance systems accumulate continuous streams that no human can watch in full~\citep{yuan2024surveillance}. In these applications, the answer to a question typically occupies seconds of content buried in hours of redundancy, so treating every frame equally wastes most computation on irrelevant content.

Vision-language models (VLMs) have responded by extending context windows and compressing visual tokens to fit more frames into a single forward pass~\citep{gemini15pro2024, zhang2024longva, xue2024longvila, shu2025videoxl, bai2025qwen2}. This strategy scales poorly with duration: a one-hour video at one frame per second yields about $1.84$ million visual tokens, exceeding most models' context windows, and the cost grows linearly with video length while the answer-relevant evidence does not. Cost-efficient long video question answering (QA) is therefore in high demand.

A promising response is the agentic \emph{harness}: a program wrapped around a frozen VLM that decides what the model observes, in what order, and when it answers~\citep{anthropic2025claudecode, tbd2026harnesssurvey}. Under this view, existing video agents~\citep{wang2024videoagent,fan2024videoagentmem,wang2025videotree,ma2025drvideo,yang2025vca,pang2025mr,zhang2025deep,lenswalk2026,lin2026videoseek} are specific harnesses, and recent ones such as VideoSeek~\citep{lin2026videoseek} and LensWalk~\citep{lenswalk2026} achieve strong accuracy with few frames. Behind these results, however, lies extensive manual engineering: experts propose, implement, test, diagnose, and revise mechanisms over many trial-and-error cycles, which demands joint expertise in video understanding and agent design and must be repeated for every new application. We ask \textbf{whether this design loop can be automated}, and answer affirmatively with \ours{}, a framework that automatically evolves agent harnesses to boost a frozen VLM on cost-efficient long video QA.

In \ours{}, a harness proposer operates inside an evolution environment: it writes candidate harnesses, executes them under a strictly enforced frame budget, analyzes the execution feedback, and revises them over iterations. Effective harness evolution and deployment, however, face three challenges. \textbf{(1) Local optima.} Vanilla harness evolution~\citep{tbd2026metaharness} revises harnesses along a single lineage and easily converges to a local optimum. We therefore organize the evolution as Monte Carlo tree search (MCTS) over a harness tree, which balances introducing new mechanisms (exploration) and refining promising harnesses (exploitation). \textbf{(2) Evaluation cost.} Fully validating every candidate on long videos takes hours and incurs high API cost, although most candidates are not competitive. We therefore propose uncertainty-aware multi-fidelity validation, which first validates new harnesses on a few questions, promotes promising ones to more questions, and penalizes accuracies measured on few questions during node selection. \textbf{(3) Diverse budgets.} Applications request different frame budgets, and the best harness under a small budget (e.g., $16$ frames) often differs from the best one under a large budget (e.g., $128$ frames). We therefore introduce a mixture-of-harness that routes each question to a harness specialized for its budget tier.

To evolve harnesses, we curate a validation set of $1{,}000$ QA pairs from existing video datasets~\citep{chen2025longvideoreason,fu2025video,wu2024longvideobench,cheng2025video,chen2025longvideoreason,zhu2026mmr}, which we will release. \ours{} sets new state-of-the-art results on LongVideoBench ($72.0\%$), Video-MME ($73.1\%$ without and $83.2\%$ with subtitles), and Video-Holmes ($60.8\%$), and outperforms VideoSeek on the same backbone by up to $11.2$ points. On the out-of-domain benchmarks Video-MMMU~\citep{hu2025videommmu} and MMVU~\citep{zhao2025mmvu}, even its lowest budget tier surpasses uniform frame sampling while using fewer than half of the frames.

In summary, our contributions are:
\begin{itemize}
    \item We propose \ours{}, a framework that automates harness design for cost-efficient long video understanding through iterative harness evolution.
    \item We design an MCTS-based evolution algorithm with uncertainty-aware multi-fidelity validation, which escapes local optima and substantially reduces the evaluation cost.
    \item We introduce a mixture-of-harness that routes each question by its frame budget.
    \item We demonstrate state-of-the-art results on LongVideoBench, Video-MME, and Video-Holmes, surpassing the hand-crafted VideoSeek by up to $11.2$ points.
\end{itemize}

\section{Related Work}
\label{sec:related}

Our work is closely related to long video understanding, agents and harnesses, and evolution algorithms. We summarize each line of work below and provide an extended discussion in Appendix~\ref{sec:appendix_related}.

\noindent\textbf{Long Video Understanding.} Existing methods scale VLMs to ingest more frames~\citep{gemini15pro2024, zhang2024longva, shu2025videoxl}, select query-relevant keyframes~\citep{tang2025aks, yu2025framevoyager, ye2025tstar}, or wrap the VLM in hand-crafted agentic harnesses that gather evidence~\citep{wang2024videoagent, wang2025videotree, zhang2025deep, lenswalk2026, lin2026videoseek}, whereas \ours{} evolves the harness automatically.

\noindent\textbf{Agents and Harnesses.} Recent work recognizes that agent capability depends heavily on the surrounding harness and optimizes harnesses for text, code, game, and, concurrently, video tasks~\citep{tbd2026harnesssurvey, tbd2026metaharness, tbd2026autoharness, xu2026videoharness, cui2026metavideoagent}, but these methods largely follow a single improvement trajectory and deploy one agent, whereas \ours{} explores a harness tree with multi-fidelity MCTS and assembles a mixture-of-harness.

\noindent\textbf{Evolution Algorithms.} LLM-driven evolution and tree search have been used to discover programs, prompts, and agentic systems~\citep{romera2024funsearch, novikov2025alphaevolve, hu2025adas, zhang2025aflow, jiang2025aide}, but most of them evaluate every candidate in full, whereas \ours{} integrates multi-fidelity evaluation~\citep{li2018hyperband} into MCTS to reduce the cost of validating harnesses on long videos.

\section{\ours{}}
\label{sec:method}

We propose \ours{}, a framework that automatically evolves agent harnesses to boost a frozen VLM for cost-efficient long video QA. We first formulate long video QA under a frame budget and the harness evolution problem in Section~\ref{sec:formulation}. Section~\ref{sec:environment} presents the evolution environment, in which candidate harnesses are written, executed, and diagnosed. Section~\ref{sec:valset} describes the validation set on which the candidate harnesses are evaluated. Section~\ref{sec:evolution} then details the multi-fidelity MCTS that navigates the harness space at a low evaluation cost. Finally, Section~\ref{sec:moh} introduces the mixture-of-harness, which routes each question to an evolved harness specialized for its frame budget.

\begin{figure}[t]
\begin{center}
\includegraphics[width=\linewidth]{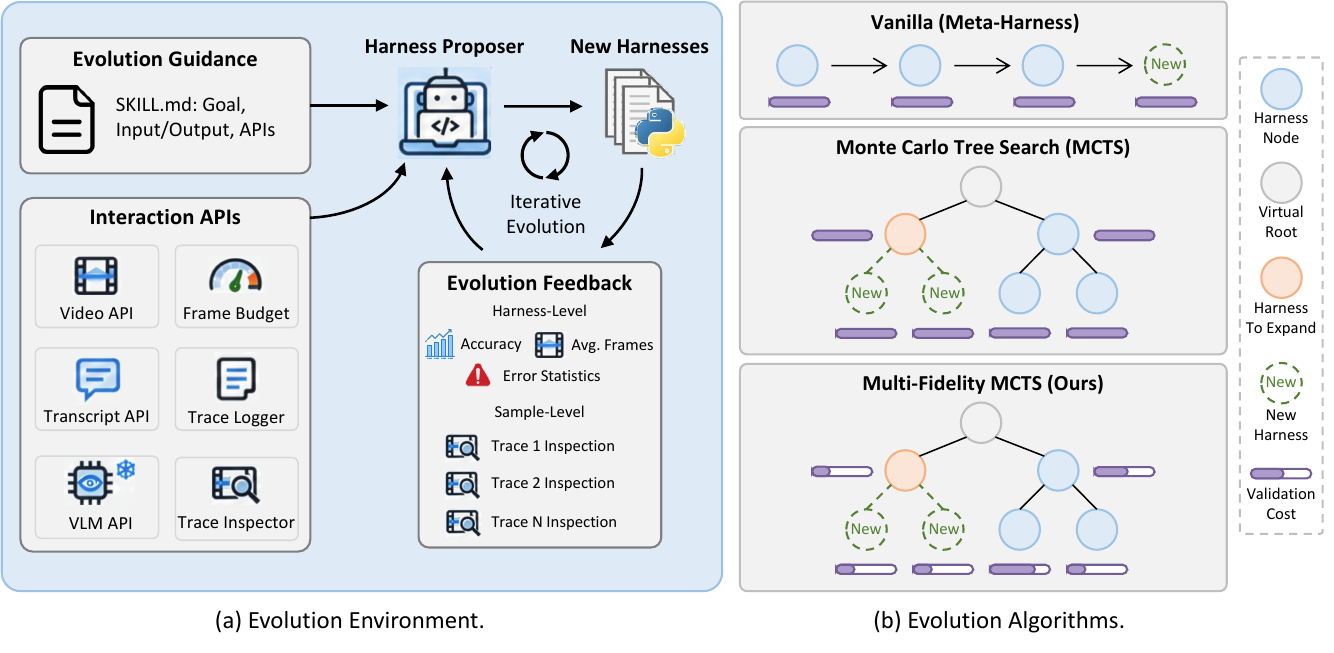}
\end{center}
\caption{Overview of harness evolution in \ours{}. Left: in the evolution environment, the harness proposer follows the evolution guidance and uses the interaction APIs to design new harnesses, whose evolution feedback is returned to the proposer for the next iteration. Right: comparison of evolution algorithms, where each circle is a harness and its bar shows the fraction of validation questions evaluated. Vanilla iterative refinement grows a single lineage and MCTS grows a tree, but both validate every harness in full; our multi-fidelity MCTS validates new harnesses on a small subset and re-validates only the promising ones.}
\label{fig:framework}
\end{figure}

\subsection{Problem Formulation}
\label{sec:formulation}

\noindent\textbf{Long Video QA under a Frame Budget.} Given a video $v$, its audio transcription $a$, a question $q$, and a frame budget $b$, the task is to design an approach that uses a frozen VLM $f$ to produce a prediction $y$ matching the ground truth $y^{\ast}$, while decoding no more than $b$ frames from $v$. As discussed in Section~\ref{sec:intro}, many video agents~\citep{wang2024videoagent,fan2024videoagentmem,wang2025videotree,ma2025drvideo,yang2025vca,pang2025mr,zhang2025deep,lenswalk2026,lin2026videoseek} build agentic frameworks around frozen VLMs that perform question-driven frame selection, which enables cost-efficient long video QA under a limited frame budget. We refer to such an agentic workflow as a \emph{harness} $h$: an executable program that orchestrates the multi-turn interaction between the video, the transcription, and the model. Running the harness yields
\begin{equation}
    y = h(v, a, q;\, f, b).
    \label{eq:harness}
\end{equation}

\noindent\textbf{Harness Evolution.} Let $\mathcal{H}$ denote the space of executable harnesses. Given a validation set $\mathcal{D}_{\mathrm{val}} = \{(v_i, a_i, q_i, y_i^{\ast})\}_{i=1}^{|\mathcal{D}_{\mathrm{val}}|}$, harness evolution seeks the harness that maximizes accuracy under the frame budget $b$:
\begin{equation}
    h^{\ast} = \argmax_{h \in \mathcal{H}} \;
    \frac{1}{|\mathcal{D}_{\mathrm{val}}|} \sum_{i=1}^{|\mathcal{D}_{\mathrm{val}}|}
    \mathbb{I}\big[ h(v_i, a_i, q_i;\, f, b) = y_i^{\ast} \big],
    \label{eq:objective}
\end{equation}
where $\mathbb{I}[\cdot]$ is the indicator function. To solve Equation~\ref{eq:objective}, we design an iterative evolution algorithm that identifies the top-performing harness $h^{\ast}$ in the harness space $\mathcal{H}$.

\subsection{Evolution Environment}
\label{sec:environment}

The evolution environment (Figure~\ref{fig:framework}, left) supports the evolution algorithm with three services, namely evolution guidance, interaction APIs, and evolution feedback, through which the harness proposer designs, executes, and diagnoses candidate harnesses.

\noindent\textbf{Evolution Guidance.} We provide the guidance to the harness proposer as a skill file (\texttt{SKILL.md})~\citep{anthropic2025skills}, which specifies the task goal, i.e., maximizing the accuracy of long video QA within the frame budget $b$, the input and output protocol of a harness, and the available APIs with their constraints. In particular, a harness is a self-contained program that realizes Equation~\ref{eq:harness} and returns its prediction together with a standardized execution trace.

\noindent\textbf{Interaction APIs.} The interaction APIs serve both the harnesses and the proposer. A harness uses the video API to sample frames within any time window at an adjustable resolution, the transcript API to retrieve the transcript by time or keyword at no frame cost, and the VLM API to converse with the frozen backbone $f$ using interleaved video frames and text. Every decoded frame is deducted from the budget $b$, and requests beyond the remaining budget are rejected. For the proposer, a trace logger records every VLM call, frame sample, and transcript read, and a trace inspector re-renders the frames a harness observed for failure analysis.

\noindent\textbf{Evolution Feedback.} After validation, the environment returns harness-level feedback, including the accuracy, the average frame usage, and error statistics, together with the sample-level trace of every question. To understand failures, the proposer dispatches trace inspectors, vision-capable subagents that examine the observed frames, diagnose causes such as missed coverage, insufficient resolution, or flawed reasoning, and return concise text reports.

\noindent\textbf{Harness Proposer.} The harness proposer $\mathcal{M}$ is instantiated as a coding agent. At each iteration, it receives the guidance, the interaction APIs, and the feedback of the harnesses evaluated so far, diagnoses their failures, and designs new harnesses that aim to improve accuracy under the frame budget. The new harnesses are then validated, and their feedback is returned to the proposer for the next iteration. Section~\ref{sec:evolution} describes how the evolution algorithm decides which harness to revise and how thoroughly to validate each new harness.

\subsection{Validation Set Construction}
\label{sec:valset}

\begin{figure}[t]
\begin{center}
\includegraphics[width=0.8\linewidth]{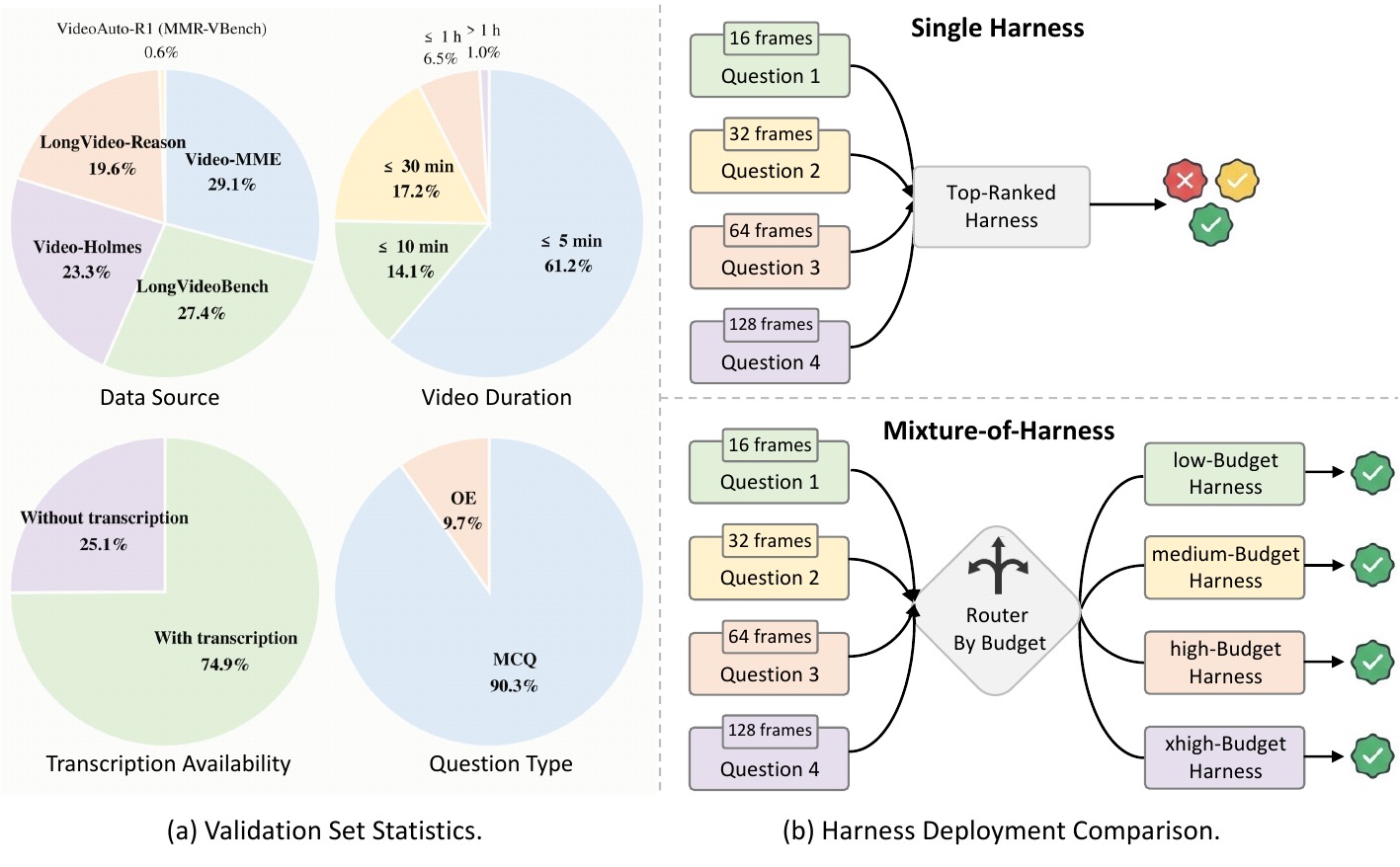}
\end{center}
\caption{Validation set statistics and mixture-of-harness. Left: distributions of data sources, video durations, transcription availability, and question types (multiple-choice, MCQ, and open-ended, OE) in the validation set. Right: a single top-ranked harness serves questions of all frame budgets but fits only some of them well (top), whereas the mixture-of-harness routes each question to the harness specialized for its budget tier (bottom).}
\label{fig:valset_moh}
\end{figure}

To facilitate harness evolution, we curate a validation set $\mathcal{D}_{\mathrm{val}}$ of $1{,}000$ QA pairs from multiple data sources~\citep{fu2025video, wu2024longvideobench, cheng2025video, chen2025longvideoreason, zhu2026mmr}. Figure~\ref{fig:valset_moh} (left) summarizes its statistics. The QA pairs mainly come from Video-MME ($29.1\%$), LongVideoBench ($27.4\%$), Video-Holmes ($23.3\%$), and LongVideo-Reason ($19.6\%$), with a small portion from VideoAuto-R1 (MMR-VBench) ($0.6\%$). We ensure that the validation set does not overlap with any test benchmark in Section~\ref{sec:experiments}. The video durations range from under $5$ minutes to over $1$ hour, which exposes the evolution to inputs of diverse lengths and encourages harnesses that handle both short and long videos. In addition, $74.9\%$ of the samples include an ASR transcription~\citep{radford2023robust} and the remaining $25.1\%$ do not, which guides the evolution to handle both cases. The questions cover both multiple-choice ($90.3\%$) and open-ended ($9.7\%$) formats. We will release the validation set to the community to support future research.

\subsection{Multi-Fidelity MCTS}
\label{sec:evolution}

Vanilla iterative refinement~\citep[Meta-Harness;][]{tbd2026metaharness} repeatedly revises an existing harness based on its evolution history and validates every candidate on the full validation set. This strategy has two limitations. First, it is greedy: since all proposals descend from the first design that works, the evolution is easily trapped in \textbf{local optima}. Second, it is slow: most of the evaluation budget is spent on candidates that turn out to be weak, resulting in a high \textbf{evaluation cost}. \ours{} instead organizes harnesses into a tree rather than a single lineage, and combines Monte Carlo tree search (MCTS), which balances exploration and exploitation, with uncertainty-aware multi-fidelity validation, which spends evaluation effort in proportion to how promising a harness is, as illustrated in Figure~\ref{fig:framework} (right) and summarized in Algorithm~\ref{alg:evolution} (Appendix~\ref{sec:appendix_alg}).

\noindent\textbf{Monte Carlo Tree Search.} We organize the harnesses designed during evolution into a harness tree $\mathcal{T}$, in which each child revises its parent. The root is a virtual node that corresponds to no harness, and every other node is a harness $h$ associated with its evolution feedback. Each iteration of MCTS consists of four stages: node selection, node expansion, validation, and backpropagation.

In node selection, the evolution algorithm descends from the root until it reaches a node that can be further expanded. At each node $h$, it moves to the child $h'$ with the highest upper confidence bound for trees (UCT) score~\citep{kocsis2006uct}:
\begin{equation}
    \mathrm{UCT}(h') = \underbrace{Q(h')}_{\text{accuracy (exploitation)}} + \underbrace{\lambda_{\mathrm{x}} \sqrt{\frac{\ln\big(N(h) + 1\big)}{N(h') + 1}}}_{\text{visit count (exploration)}},
    \label{eq:uct}
\end{equation}
where $Q(h')$ is the highest accuracy among the harnesses in the subtree rooted at $h'$, $N(\cdot)$ is the visit count of a node, i.e., the number of harnesses evaluated in its subtree, and $\lambda_{\mathrm{x}}$ is the exploration weight. The first term favors branches that contain accurate harnesses, and the second term favors branches that have rarely been visited.

In node expansion, the harness proposer designs new harnesses as children of the selected node $h$, conditioned on the current harness tree $\mathcal{T}$, which contains all evaluated harnesses and their feedback. We denote this operation by $\mathcal{M}(\mathcal{T})$. If the selected node is the virtual root, the proposer designs harnesses whose core mechanisms differ from all harnesses in the tree, which encourages exploration. Otherwise, it revises the selected harness, which encourages exploitation.

In validation, each new harness is executed on the validation set $\mathcal{D}_{\mathrm{val}}$ under the frame budget $b$, which yields its accuracy $\mathrm{acc}(h)$ and its feedback.

In backpropagation, every ancestor of a new node increments its visit count $N(\cdot)$ and updates $Q(\cdot)$ if the new harness is more accurate than all harnesses in its subtree. We adopt this max-style backup rather than the usual average because evolution seeks the single best harness.

\noindent\textbf{Uncertainty-Aware Multi-Fidelity Validation.} To reduce the evaluation cost, we propose uncertainty-aware multi-fidelity validation, inspired by multi-fidelity hyperparameter optimization~\citep{jamieson2016nonstochastic, li2018hyperband, falkner2018bohb}. We define $L$ rungs $r_1 < \dots < r_L = |\mathcal{D}_{\mathrm{val}}|$, where rung $r_\ell$ covers the first $r_\ell$ questions of $\mathcal{D}_{\mathrm{val}}$. Let $\mathrm{acc}_r(h)$ denote the accuracy of $h$ on the first $r$ questions, and let $r(h)$ denote the number of questions on which harness $h$ has been validated. Each new harness is first validated only at rung $r_1$, which substantially accelerates evolution and reduces its cost. However, an accuracy measured on a few questions is noisy: its standard error can reach $1/(2\sqrt{r})$ (Appendix~\ref{sec:appendix_se}), so it is an unreliable estimate of the true accuracy of $h$. We address this uncertainty with two designs.

First, we add a promotion stage at the end of each iteration. Among the harnesses that have not yet been validated on the full validation set, we select the one with the highest accuracy,
\begin{equation}
    h^{\dagger} = \argmax_{h:\; r(h) < |\mathcal{D}_{\mathrm{val}}|} \mathrm{acc}_{r(h)}(h),
    \label{eq:promotion}
\end{equation}
and re-validate it at the next rung. We promote the current leader because the reliability of the best harness's accuracy matters most: if its accuracy holds, the promotion confirms a promising harness; if it drops, the lead is attributed to sampling noise. Since the rungs are nested, a promotion only evaluates the questions that the harness has not yet answered.

Second, we add a low-fidelity penalty to node selection, which discounts a high accuracy that is measured on only a few questions. Specifically, we define the penalized accuracy of a harness as a lower confidence bound on its accuracy:
\begin{equation}
    \underline{\mathrm{acc}}(h) = \mathrm{acc}_{r(h)}(h) - \lambda_{\mathrm{p}} \cdot \frac{1}{2\sqrt{r(h)}},
    \label{eq:value}
\end{equation}
where $\frac{1}{2\sqrt{r(h)}}$ is the upper bound on the standard error of the accuracy measured on $r(h)$ questions (Appendix~\ref{sec:appendix_se}), and $\lambda_{\mathrm{p}}$ controls the strength of the penalty. For example, with $\lambda_{\mathrm{p}} = 1.5$, the penalty is $7.5$ percentage points on $100$ questions but only $2.4$ points on $1{,}000$ questions, so an accuracy of $76\%$ on $100$ questions does not outrank an accuracy of $72\%$ on $1{,}000$ questions. We then use the penalized accuracy in the exploitation term of Equation~\ref{eq:uct}, i.e., $Q(h')$ becomes the highest penalized accuracy among the harnesses in the subtree rooted at $h'$. As a result, a branch whose best harness has only been validated on a few questions attracts less exploitation. After the evolution, all harnesses are ranked by their penalized accuracy.

\subsection{Mixture-of-Harness}
\label{sec:moh}

Multi-fidelity MCTS returns a tree of harnesses ranked under a single frame budget $b$. In practice, however, questions arrive with different budget requests: a latency- or cost-sensitive application may afford only a few frames per question, whereas offline analysis may afford over a hundred. Across this range, the relative performance of the evolved harnesses varies with the budget. Harnesses that commit to an answer after a brief survey of the video use a small budget efficiently but benefit little from additional frames, whereas harnesses that continue to gather evidence perform well under a generous budget but degrade substantially when the budget is reduced. As a result, the best harness under a low budget often differs from that under a high budget, so a single top-ranked harness is suboptimal for part of the budget range. We therefore propose a mixture-of-harness (MoH), which routes each question to a harness specialized for its budget (Figure~\ref{fig:valset_moh}, right).

\noindent\textbf{Member Selection.} We define a set of budget tiers $\mathcal{K} = \{\mathrm{low}, \mathrm{medium}, \mathrm{high}, \mathrm{xhigh}\}$, where each tier $k \in \mathcal{K}$ corresponds to a range of frame budgets (Appendix~\ref{sec:appendix_impl}). After evolution, we reuse the harness proposer $\mathcal{M}$ to analyze the evolved harnesses. For each harness $h$ in $\mathcal{T}$, $\mathcal{M}$ inspects its code together with the feedback collected on the validation set during evolution, and characterizes its behavior pattern, including its frame allocation over the video, its stopping behavior (answering early or exhausting the budget), and its reliance on the transcript for evidence localization. Based on this analysis, $\mathcal{M}$ assigns to each tier the harness that best suits its budget range:
\begin{equation}
    \{h_{k}\}_{k \in \mathcal{K}} = \mathcal{M}(\mathcal{T}, \mathcal{K}).
    \label{eq:member}
\end{equation}
Member selection relies only on the harness code and the feedback collected during evolution, so it requires no additional harness executions. The selected harnesses may come from different lineages of the harness tree, and the same harness may serve several tiers.

\noindent\textbf{Routing by Budget.} At inference, each question $q$ arrives with a requested frame budget $b$, for example chosen to meet a latency or cost target. MoH identifies the tier $k \in \mathcal{K}$ whose range contains $b$ and routes the question to the harness of that tier:
\begin{equation}
    y = h_{k}(v, a, q;\, f, b).
    \label{eq:moh}
\end{equation}
Since routing depends only on the budget rather than the video or question, it requires no learned router and adds no runtime cost. Section~\ref{sec:ablation} reports its accuracy at each budget tier.

\section{Experiments}
\label{sec:experiments}

\begin{table}[t]
\caption{Performance on long video understanding benchmarks: LongVideoBench, VideoMME, and LVBench. For VideoMME and LVBench, accuracy is reported both with and without subtitles. For our approach, \ours{}, we report results under low, medium, high, and extra-high (xhigh) inference frame budgets. \#Frames indicates the number of frames each method processes. \textbf{Bold} indicates the best result and \underline{underline} the second best. }
\label{tab:sota_long}
\begin{center}
\setlength{\tabcolsep}{4pt}
\begin{adjustbox}{width=1\textwidth}\setlength\tabcolsep{5pt}
\begin{tabular}{llcccccccccc}
\toprule
\multirow{2}{*}{\textbf{Method}} &
\multirow{2}{*}{\textbf{Model}} &
\multicolumn{2}{c}{\textbf{LongVideoBench}} &
\multicolumn{2}{c}{\textbf{VideoMME} (w/o sub)} &
\multicolumn{2}{c}{\textbf{VideoMME} (w/ sub)} &
\multicolumn{2}{c}{\textbf{LVBench} (w/o sub)} &
\multicolumn{2}{c}{\textbf{LVBench} (w/ sub)} \\
\cmidrule(lr){3-4}  \cmidrule(lr){5-6} \cmidrule(lr){7-8}  \cmidrule(lr){9-10} \cmidrule(lr){11-12}
& & \#Frames & Long (val) & \#Frames & Long & \#Frames & Long & \#Frames & Test & \#Frames & Test \\
\hline
\multicolumn{12}{>{\columncolor[gray]{.88}}l}{\emph{Large Multimodal Models}} \\
Qwen-2.5-VL-72B~\citep{bai2025qwen2}        & & -    & -    & 256   & 53.2 & 256   & 64.4 & 768  & 47.3 & -   & -    \\
GPT-4o~\citep{gpt4o2024}                & & 256  & 60.9 & 384   & 65.3 & 384   & 72.1 & 384  & 30.8 & -   & -    \\
Gemini-1.5-Pro~\citep{gemini15pro2024}       & & 256  & 58.6 & 1,233  & 67.4 & 1,233  & 77.4 & 3,600 & 33.1 & -   & -    \\
Gemini-2.0-Flash~\citep{gemini20flash2024}      & & 256  & 45.7 & 1,233  & 63.0 & -  & -    & 4,037 & 48.3 & -   & -    \\
GPT-5~\citep{gpt52025}          & & 384  & 64.5 & 384   & 67.9 & 384   & 78.1 & 384  & 60.1 & 384 & 66.5 \\
Qwen-3.6-27B~\citep{qwen2026qwen36} & & 64  & 65.6 & 64 & \underline{68.0} &  64 & 79.8 & 64  & 55.5  & 64 & 63.5 \\
\hline
\multicolumn{12}{>{\columncolor[gray]{.88}}l}{\emph{Video Agentic Models}} \\
VideoAgent~\citep{wang2024videoagent}            & GPT-4+LaViLa & -    & -    & 24.6  & 46.4 & -     & -    & 25.5  & 29.3 & -   & -    \\
VideoTree~\citep{wang2025videotree}             & GPT-4+LaViLa & -    & -    & 98.0    & 53.1 & -     & -    & 103.2 & 28.8 & -   & -    \\
DrVideo~\citep{ma2025drvideo}               & GPT-4+LaViLa & -     & -    & 493.2 & 51.7 & 493.2 & 71.7 & -     & -    & -   & -    \\
VCA~\citep{yang2025vca}                   & GPT-4o & -     & -    & 18.1  & 54.2 & -     & -    & 20.0    & 41.3 & -   & -    \\
MR.~Video~\citep{pang2025mr}            & GPT-4o+Gemini-2.0-Flash & 2,816 & 61.6 & 4,932  & 61.8 & 4,932  & -    & 8,074  & 60.8 & -   & -    \\
DVD~\citep{zhang2025deep}  & GPT-o3+GPT-4.1 & 2,816 & 68.6 & 4,932 & 67.3 & 4,932  & -    & 8,074  & \textbf{74.2} & 8,074 & \textbf{76.0} \\
VideoSeek~\citep{lin2026videoseek}     & GPT-5 & \textcolor{gray!50}{29.6} & \textcolor{gray!50}{73.5} & \textcolor{gray!50}{60.9} & \textcolor{gray!50}{70.1} & \textcolor{gray!50}{15.9}  & \textcolor{gray!50}{81.2} & \textcolor{gray!50}{92.3}    & \textcolor{gray!50}{68.4} & \textcolor{gray!50}{27.2} & \textcolor{gray!50}{76.7} \\
VideoSeek~\citep{lin2026videoseek}     & Qwen-3.6-27B & 57.5 & 67.9 & 123.0 & 65.0 & 17.6 & 79.6 & 111.5 & 54.7 & 45.9 & 63.3  \\
\hline
\multicolumn{12}{>{\columncolor{myblue!15}}l}{\emph{\ours~(Ours)}} \\
low & Qwen-3.6-27B & 23 & 67.7 & 22 & 62.0 & 18 & 78.7 & 25 & 49.8 & 22 & 62.6 \\
medium & Qwen-3.6-27B & 44 & 69.1 & 43 & 65.9 & 36 & 80.3 & 50 & 54.7 & 43 & 63.5 \\
high & Qwen-3.6-27B& 68 & \underline{71.1} & 67 & 67.4 & 66 & \underline{82.8} & 70 & 56.0 & 69 & 66.1 \\
xhigh & Qwen-3.6-27B & 126 & \textbf{72.0} & 126 & \textbf{73.1} & 126 & \textbf{83.2} & 126 & \underline{61.1} & 126 & \underline{66.7} \\
\bottomrule
\end{tabular}
\end{adjustbox}
\end{center}
\end{table}
\subsection{Experimental Details}
\label{sec:exp_details}

\noindent\textbf{Benchmarks.} Following the evaluation protocol of VideoSeek~\citep{lin2026videoseek}, we evaluate on six long video benchmarks: LVBench~\citep{wang2025lvbench} ($1{,}549$ questions over $103$ hour-long videos), the long subset of Video-MME~\citep{fu2025video} ($900$ questions, average duration $41$ minutes), the long-duration validation split of LongVideoBench~\citep{wu2024longvideobench} ($564$ questions, $900$ to $3{,}600$ seconds), Video-Holmes~\citep{cheng2025video} ($1{,}837$ questions over $270$ suspense films), Video-MMMU~\citep{hu2025videommmu} ($900$ questions on professional lecture videos), and MMVU~\citep{zhao2025mmvu} ($1{,}000$ expert-level questions). For LVBench and Video-MME we report both with- and without-subtitle settings where applicable.

\noindent\textbf{Baselines.} We compare against two groups of approaches: large multimodal models and video agentic models. Large multimodal models process uniformly sampled frames in a single pass: Qwen-2.5-VL-72B~\citep{bai2025qwen2}, GPT-4o~\citep{gpt4o2024}, Gemini 1.5 Pro~\citep{gemini15pro2024}, Gemini 2.0 Flash~\citep{gemini20flash2024}, GPT-5~\citep{gpt52025}, Qwen-3.6-27B~\citep{qwen2026qwen36}. Video agentic models actively gather evidence: VideoAgent~\citep{wang2024videoagent}, VideoTree~\citep{wang2025videotree}, DrVideo~\citep{ma2025drvideo}, VCA~\citep{yang2025vca}, MR.~Video~\citep{pang2025mr}, DVD~\citep{zhang2025deep}, and VideoSeek~\citep{lin2026videoseek} (current state-of-the-art approach). On Video-Holmes we additionally include some reasoning-enhanced video models~\citep{chen2025exploring,li2025videochat,feng2025video}. For a fair comparison, we reproduce VideoSeek's results with Qwen-3.6-27B to isolate the effect of backbone models.

\noindent\textbf{Implementation Details.} We provide the implementation details, including the harness proposer, the evolution hyperparameters, and the budget tiers of the mixture-of-harness, in Appendix~\ref{sec:appendix_impl}.

\subsection{Comparison with State-of-the-Art Methods}
\label{sec:sota}
Table~\ref{tab:sota_long} reports the comparison on LongVideoBench, Video-MME, and LVBench (with and without subtitles). At its highest budget (xhigh, about $128$ frames), \ours{} reaches $72.0\%$ on LongVideoBench, $73.1\%$ on Video-MME without subtitles, and $83.2\%$ on Video-MME with subtitles, outperforming all existing baselines and setting a new state of the art on all three settings. On both Video-MME settings, accuracy with subtitles is consistently higher than without, showing that \ours{} makes effective use of the transcribed audio content to assist long video understanding. On LVBench, in contrast, \ours{} falls behind DVD in both the with-subtitle and without-subtitle settings, because DVD processes an order of magnitude more frames per question ($8{,}074$ against \ours{}'s $126$). VideoSeek is the state-of-the-art video agentic model and thus the most direct baseline for \ours{}; against VideoSeek with the same Qwen-3.6-27B backbone, \ours{} shows a clear advantage: on LVBench without subtitles, VideoSeek reaches $54.7\%$ with $111.5$ frames, while \ours{} reaches $61.1\%$ with $126$ frames and surpasses VideoSeek, $56.0\%$, with only $70$ frames; on LVBench with subtitles, \ours{} reaches $63.5\%$ with $43$ frames against VideoSeek's $63.3\%$ with $45.9$ frames. \ours{} at its high tier also generally outperforms the Qwen-3.6-27B uniform-sampling baseline across the five settings.

Table~\ref{tab:sota_holmes} shows the results on Video-Holmes, which requires multi-step reasoning over suspense films. \ours{} obtains $60.8\%$ overall at its xhigh tier, surpassing VideoSeek with the same backbone by $11.2$ points and the uniform-sampling baseline by $1.3$ points, with the largest gains over VideoSeek on the Physical Anomaly Reasoning ($+16.5$) and Temporal Causal Inference ($+14.7$) categories. \ours also outperforms Qwen-3.6-27B with similar frame budgets ($59.3\%$ against $58.9\%$).

\begin{table}[t]
\caption{Comparison on Video-Holmes. The seven reasoning categories are: SR is social reasoning, IMC is intention and motive chaining, TCI is temporal causal inference, TA is timeline analysis, MHR is multimodal hint reasoning, PAR is physical anomaly reasoning, and CTI is core theme inference. For our approach, \ours{}, we report results under low, medium, high, and extra-high (xhigh) inference frame budgets. \#Frames denotes the frame usage. \textbf{Bold} marks the best performance, and \underline{underline} marks the second-best. }
\label{tab:sota_holmes}
\begin{center}
\begin{adjustbox}{width=1\textwidth}\setlength\tabcolsep{10pt}
\begin{tabular}{lccccccccc}
\toprule
\textbf{Method} & \textbf{\#Frames} & \textbf{SR} & \textbf{IMC} & \textbf{TCI} & \textbf{TA} & \textbf{MHR} & \textbf{PAR} & \textbf{CTI} & \textbf{Overall} \\
\hline
\multicolumn{10}{>{\columncolor[gray]{.88}}l}{\emph{Large Multimodal Models}} \\
Qwen-2.5-VL-32B~\citep{bai2025qwen2} & 32   & 43.2  & 44.2  & 31.5  & 51.0 & 36.4  & 31.4  & 32.2  & 38.4  \\
SEED-Bench-R1~\citep{chen2025exploring} & 32   & 42.8  & 35.1  & 25.6  & 40.5 & 29.2  & 29.9  & 32.6  & 33.5  \\
VideoChat-R1~\citep{li2025videochat}   & 32   & 42.1  & 38.8  & 24.5  & 39.5 & 29.5  & 27.8  & 29.3  & 33.0  \\
Video-R1~\citep{feng2025video}       & 32   & 48.6  & 41.7  & 28.9  & 34.5 & 31.0  & 33.5  & 35.9  & 36.5  \\
GPT-4o~\citep{gpt4o2024}         & 32   & 50.0  & 49.6  & 38.8  & 30.0 & 44.0  & 39.2  & 37.0  & 42.0  \\
Gemini~1.5~Pro~\citep{gemini15pro2024} & 185.1   & 52.1  & 48.2  & 34.4  & 26.0 & 39.2  & 46.4  & 38.9  & 41.2  \\
Gemini~2.5~Pro~\citep{gemini25pro2025} & 185.1   & 46.6  & 49.3 & 46.9  & 53.0 & 40.1  & 44.3  & 37.4  & 45.0  \\
Gemini~2.0~Flash~\citep{gemini20flash2024} & 185.1 & 41.8  & 33.7  & 23.1  & 20.5 & 30.1  & 26.8  & 33.7  & 30.6  \\
Gemini~2.0~Flash~Thinking~\citep{gemini20flashthinking2025} & 185.1 & 43.4 & 46.9 & 43.1 & 51.0 & 37.9 & 43.6 & 39.3 & 43.1 \\
GPT-5~\citep{gpt52025} (Base)   & 384  & 47.2 & 43.4  & 40.6 & 53.5 & 46.3  & 38.1 & 39.6 & 44.1  \\
Qwen-3.6-27B~\citep{qwen2026qwen36} & 63.8  & \underline{68.5} & 63.0 & 58.6 & 59.0 & \underline{57.5} & 52.6 & 50.7 & 58.9 \\
\hline
\multicolumn{10}{>{\columncolor[gray]{.88}}l}{\emph{Video Agentic Models}} \\
VideoSeek~\citep{lin2026videoseek} (GPT-5) & \textcolor{gray!50}{42.7} & \textcolor{gray!50}{56.1} & \textcolor{gray!50}{43.8} & \textcolor{gray!50}{45.0} & \textcolor{gray!50}{54.5} & \textcolor{gray!50}{46.6} & \textcolor{gray!50}{43.3} & \textcolor{gray!50}{41.8} & \textcolor{gray!50}{47.3} \\
VideoSeek~\citep{lin2026videoseek} (Qwen 3.6 27B) & 68.7 & 60.3 & 53.3 & 44.3 & 49.0 & 46.1 & 45.4 & 47.4 & 49.6 \\
\hline
\multicolumn{10}{>{\columncolor{myblue!15}}l}{\emph{\ours~(Ours)}} \\
low & 21.4 & 67.5 & 60.1 & 51.3 & 48.5 & 55.4 & 52.6 & 49.3 & 55.5 \\
medium & 43.3 & 62.0 & \textbf{66.7} & \underline{59.0} & 54.0 & \textbf{57.5} & 58.3 & 52.2 & 58.7 \\
high & 82.0 & 67.5 & 62.0 & 56.0 & \textbf{62.5} & 55.7 & \underline{60.0} & \underline{52.6} & \underline{59.3} \\
xhigh & 126.0 & \textbf{69.2} & \underline{65.6} & \textbf{59.0} & 60.0 & 56.0 & \textbf{61.9} & \textbf{54.4} & \textbf{60.8} \\
\bottomrule
\end{tabular}
\end{adjustbox}
\end{center}
\end{table}

\subsection{Ablation Studies}
\label{sec:ablation}

We ablate the two central designs of \ours{}, the evolution strategy and the mixture-of-harness, on Video-Holmes, keeping the backbone and the validation set fixed.

\begin{table}[t]
\begin{minipage}[t]{0.53\linewidth}
\centering
\caption{Ablation studies on Video-Holmes. Top: evolution strategies, where Evolution Time is the total evolution time and Val.\ Samples is the total number of validation samples consumed during evolution. Bottom: mixture-of-harness (MoH) versus the top-1 harness from evolution at each of the four budget tiers (low, medium, high, and xhigh).}
\label{tab:ablation}
\begin{adjustbox}{width=\linewidth}
\begin{tabular}{lcccc}
\toprule
\textbf{Method} & \textbf{Evolution Time (h)} & \textbf{Val.\ Samples (K)} & \textbf{\#Frames} & \textbf{Video-Holmes}   \\
\midrule
Qwen-3.6-27B~\citep{qwen2026qwen36} & - & - & 63.8 & 58.9 \\
Vanilla~\citep{tbd2026metaharness} & 75.5 & 40.0 & 64.0 & 57.3 \\
MCTS                        & 71.9 & 40.0 & 64.0 & 58.4  \\
MCTS + MF  & 34.9 & 7.8 & 56.8 & 59.4 \\
\midrule
MCTS + MF + MoH (low) & - & - & 21.4 & 55.5 \\
MCTS + MF + Top1 (low) & - & - & 21.2 & 55.2\\
MCTS + MF + MoH (medium) & - & - & 43.3 & 58.7 \\
MCTS + MF + Top1 (medium) & - & - & 40.1 & 57.5\\
MCTS + MF + MoH (high) & - & - & 56.8 & 59.4 \\
MCTS + MF + Top1 (high) & - & - & 56.8 & 59.4\\
MCTS + MF + MoH (xhigh) & - & - & 126.0 & 60.8 \\
MCTS + MF + Top1 (xhigh) & - & - & 82.0 & 59.3 \\
\bottomrule
\end{tabular}
\end{adjustbox}
\end{minipage}
\hfill
\begin{minipage}[t]{0.44\linewidth}
\centering
\caption{Comparison on knowledge-intensive, out-of-domain video benchmarks, Video-MMMU and MMVU.}
\label{tab:sota_knowledge}
\begin{adjustbox}{width=\linewidth}
\begin{tabular}{lcccc}
\toprule
\multirow{2}{*}{\textbf{Method}} & \multicolumn{2}{c}{\textbf{Video-MMMU}} & \multicolumn{2}{c}{\textbf{MMVU}} \\
\cmidrule(lr){2-3} \cmidrule(lr){4-5}
& \#Frames & Acc. & \#Frames & Acc. \\
\midrule
Qwen-3.6-27B~\citep{qwen2026qwen36} & 63 & 85.3 & 50 & 74.4 \\
VideoSeek~\citep{lin2026videoseek} (Qwen 3.6 27B) & 16.8 & 79.1 & 76.0 & 69.0 \\
\ours~(Ours, low) & 20 & \textbf{85.8} & 22 & \textbf{76.8} \\
\bottomrule
\end{tabular}
\end{adjustbox}

\vspace{1em}
\includegraphics[width=\linewidth]{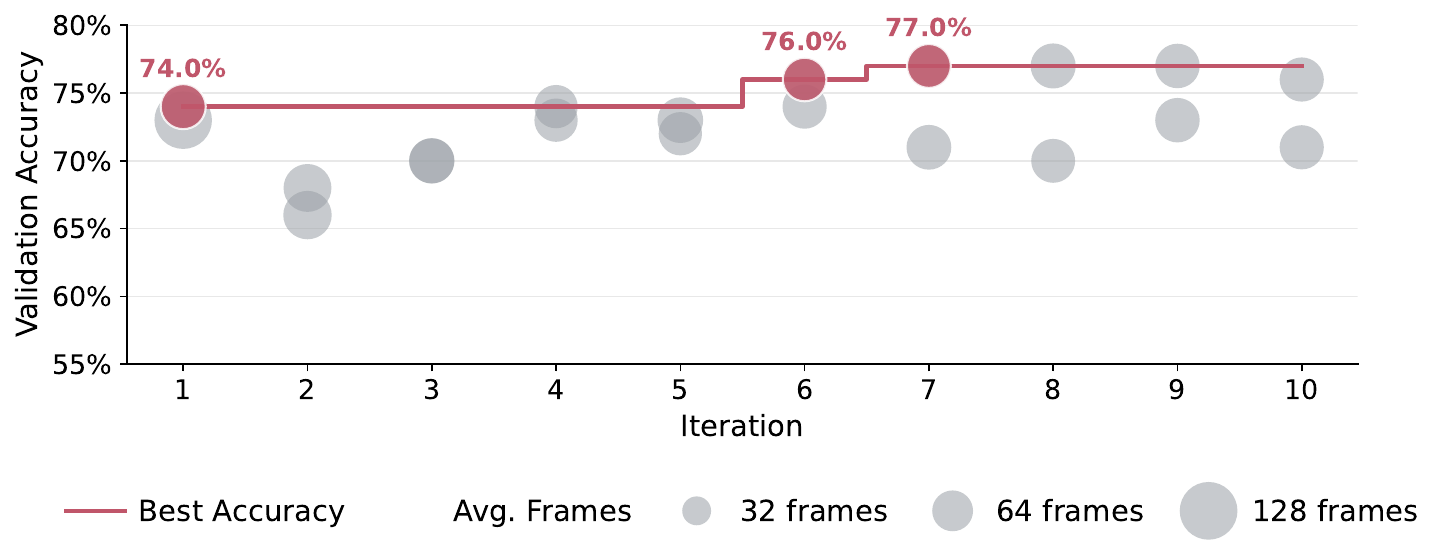}
\vspace{-2em}
\captionof{figure}{Validation accuracy over evolution iterations. Each circle is a harness designed at that iteration, with its size indicating the average number of frames used, and the red line tracks the best accuracy so far.}
\label{fig:vis}
\end{minipage}
\end{table}

\noindent\textbf{Evolution Strategy.} The top half of Table~\ref{tab:ablation} compares three evolution strategies: vanilla iterative refinement following Meta-Harness~\citep{tbd2026metaharness}, which revises harnesses along a single lineage and validates every candidate on the full validation set; MCTS with full validation; and MCTS with uncertainty-aware multi-fidelity validation (MCTS + MF). Confined to a single lineage, vanilla iterative refinement reaches only $57.3\%$ after $75.5$h of evolution, below the uniform-sampling baseline ($58.9\%$). MCTS improves the accuracy to $58.4\%$ by exploring multiple lineages, but it still validates every candidate on all $1{,}000$ questions and requires $71.9$h. MCTS + MF reduces the validation samples by $5.1\times$ and the evolution time by $2.1\times$ relative to MCTS, while achieving the highest accuracy of $59.4\%$ with fewer frames ($56.8$ versus $64.0$).

\noindent\textbf{Mixture-of-Harness.} The bottom half of Table~\ref{tab:ablation} compares MoH with the top-1 harness from evolution, which is applied to every budget tier. MoH outperforms the top-1 harness at the low ($55.5\%$ versus $55.2\%$), medium ($58.7\%$ versus $57.5\%$), and xhigh ($60.8\%$ versus $59.3\%$) tiers. At the high tier, MoH selects the top-1 harness itself, so the two results are identical ($59.4\%$). Notably, at the xhigh tier, the top-1 harness uses only $82.0$ frames on average, whereas the MoH member uses $126.0$, showing that a single harness may fail to exploit a larger budget.

\subsection{Generalization to Out-of-Domain Benchmarks}
\label{sec:generalization}

The validation set of Section~\ref{sec:valset} and the benchmarks of Tables~\ref{tab:sota_long} and~\ref{tab:sota_holmes} all concern long video understanding. To test whether the evolved harnesses generalize beyond this domain, we report results on Video-MMMU and MMVU (Table~\ref{tab:sota_knowledge}), two knowledge-intensive benchmarks. Even the low tier of \ours{}, which uses $20$ and $22$ frames, respectively, matches or exceeds the uniform-sampling baseline ($+0.5$ and $+2.4$ points) while using less than a third and less than half of its frames, and outperforms the same-backbone VideoSeek by $6.7$ and $7.8$ points.

\subsection{Evolution Dynamics}
\label{sec:evolution_dynamics}

Figure~\ref{fig:vis} shows the validation accuracy of the harnesses designed over the first $10$ iterations of evolution. The best accuracy improves from $74.0\%$ at the first iteration to $76.0\%$ at the sixth and $77.0\%$ at the seventh iteration, indicating that the evolution algorithm progressively discovers better harnesses from execution feedback. From the third iteration on, every new harness reaches at least $70\%$, suggesting that the proposer accumulates useful design knowledge across iterations.

\section{Conclusion}
\label{sec:conclusion}

We presented \ours{}, a framework that automatically evolves agent harnesses for cost-efficient long video understanding. \ours{} organizes evolution as Monte Carlo tree search with uncertainty-aware multi-fidelity validation, which escapes local optima at a low evaluation cost, and deploys a mixture-of-harness that routes each question to a budget-specialized harness. Experiments show that \ours{} achieves state-of-the-art results on three long video benchmarks.

\newpage
\subsection*{AI use statement}

In this work, we used generative AI tools (Claude Code) to implement methods: Claude Code serves as the harness proposer, i.e., the LLM backbone that proposes and edits agent harnesses within \ours{}, and is therefore a core component of our experimental method rather than only a writing aid. We have not used generative AI tools for the other tasks with required disclosure (generating synthetic datasets, developing theoretical models or conceptual frameworks, formulating mathematical claims, proving mathematical claims, assisting with translation, cleaning/reformatting datasets, supporting qualitative or thematic data analysis, or interpreting results); these are not applicable to this work. Additionally, we used generative AI tools for tasks with recommended disclosure: creating and editing scientific figures from our experimental results, and editing the paper to fix grammar and rephrase less-polished passages. We have reviewed all AI-assisted work: all LLM-proposed harnesses were evaluated through our experimental pipeline and validated against held-out data as described in the paper, and all AI-edited text and figures were checked by the authors for accuracy and consistency with the underlying results. We take responsibility for the final content of this work, including text, claims, and artifacts produced with the aid of generative AI.

\bibliography{iclr2027_conference}
\bibliographystyle{iclr2027_conference}

\appendix
\newpage
\section{Extended Related Work}
\label{sec:appendix_related}

\noindent\textbf{Long Video Understanding.} Existing methods either scale VLMs to the video by extending context windows and compressing visual tokens~\citep{gemini15pro2024, zhang2024longva, xue2024longvila, shu2025videoxl, bai2025qwen2, song2024moviechat}, or reduce the input by selecting query-relevant keyframes~\citep{tang2025aks, yu2025framevoyager, ye2025tstar}. Another line of work strengthens video reasoning through reinforcement fine-tuning~\citep{feng2025video, li2025videochat, chen2025exploring}. Closest to our setting, agentic approaches wrap the VLM in a reasoning loop that plans which evidence to gather~\citep{wang2024videoagent, fan2024videoagentmem, wang2025videotree, ma2025drvideo, yang2025vca, pang2025mr, zhang2025deep}. Among them, VideoSeek~\citep{lin2026videoseek} equips a reasoning model with a multi-granular seeking toolkit, and LensWalk~\citep{lenswalk2026} lets the model plan its own observation scope and sampling density. Although these agents achieve strong accuracy with few frames, each is a fixed design hand-crafted by experts around a particular model. In contrast, \ours{} evolves the harness automatically and adapts it to the backbone and the frame budget.

\noindent\textbf{Agents and Harnesses.} Large language model agents interleave reasoning with actions and tool calls~\citep{yao2023react, shinn2023reflexion, schick2023toolformer}, and much of their capability is determined by the surrounding \emph{harness}, the runtime that shapes observations, context, control flow, and verification~\citep{yang2024sweagent, anthropic2025claudecode, tbd2026harnesssurvey}. Recent work engineers this layer explicitly~\citep{tbd2026harnessx, tbd2026llmascode, tbd2026rah} or optimizes it automatically~\citep{tbd2026autoharness}. Among them, Meta-Harness~\citep{tbd2026metaharness} optimizes harnesses for text tasks with an outer-loop coding agent. Concurrent work brings harness optimization to video. VideoHarness-RSI~\citep{xu2026videoharness} recursively improves context-construction programs for a frozen VLM. MetaVideoAgent~\citep{cui2026metavideoagent} evolves a modular video agent for a target video distribution: it diagnoses recurring failures against ground-truth evidence, updates only the responsible module in each iteration, and tests modifications on minimal validation tasks derived from these failures. Both largely follow a single improvement trajectory and deploy one evolved agent. In contrast, \ours{} explores a harness tree with multi-fidelity MCTS and assembles harnesses specialized for different frame budgets into a mixture-of-harness.

\noindent\textbf{Evolution Algorithms.} With LLMs as mutation operators, evolution algorithms discover programs~\citep{romera2024funsearch, novikov2025alphaevolve}, optimize prompts~\citep{yang2024opro, guo2024evoprompt, fernando2024promptbreeder}, and design agentic systems. ADAS~\citep{hu2025adas} programs new agents in code space with a meta agent, AFlow~\citep{zhang2025aflow} applies MCTS to workflow graphs, and related efforts search modular agent spaces or pursue self-referential improvement~\citep{shang2025agentsquare, zhang2025maas, yin2025godelagent, zhang2025dgm}. Tree search over code also drives automated machine-learning engineering~\citep{jiang2025aide}. However, most of these methods evaluate every candidate in full on tasks where evaluation is cheap. Since evaluation on long videos is expensive, \ours{} combines MCTS~\citep{kocsis2006uct} with multi-fidelity evaluation in the spirit of successive halving and Hyperband~\citep{jamieson2016nonstochastic, li2018hyperband, falkner2018bohb}, where every arm is a harness program. Finally, our mixture-of-harness is related to mixture-of-experts~\citep{jacobs1991adaptive, shazeer2017moe} and LLM routing~\citep{chen2023frugalgpt, ong2025routellm}, which select experts or models based on the input content. In contrast, the mixture-of-harness selects among harnesses built around one frozen VLM based on the requested frame budget.

\section{Algorithm of Uncertainty-Aware Multi-Fidelity MCTS}
\label{sec:appendix_alg}

Algorithm~\ref{alg:evolution} summarizes the evolution procedure of Section~\ref{sec:evolution}. The harness tree is initialized with a virtual root. In each iteration, node selection descends from the root by the UCT score in Equation~\ref{eq:uct} until it reaches a node that can be further expanded. The harness proposer then expands the selected node with new harnesses, each of which is validated on the first rung and scored by its penalized accuracy in Equation~\ref{eq:value}. Backpropagation updates the visit counts and the exploitation terms of the ancestors of the new harnesses. Finally, the promotion stage selects the most accurate harness that has not been validated on the full validation set (Equation~\ref{eq:promotion}), re-validates it at the next rung, and updates its penalized accuracy together with the exploitation terms of its ancestors. After all iterations, the harnesses in the tree are ranked by their penalized accuracy.

\begin{algorithm}[ht]
\caption{MCTS with uncertainty-aware multi-fidelity validation in \ours{}}
\label{alg:evolution}
\begin{algorithmic}[1]
\Require validation set $\mathcal{D}_{\mathrm{val}}$; frozen VLM $f$; frame budget $b$; harness proposer $\mathcal{M}$; number of iterations $I$; rungs $r_1 < \dots < r_L = |\mathcal{D}_{\mathrm{val}}|$; exploration weight $\lambda_{\mathrm{x}}$; penalty coefficient $\lambda_{\mathrm{p}}$
\Ensure harnesses in the harness tree $\mathcal{T}$ ranked by penalized accuracy $\underline{\mathrm{acc}}$
\State initialize $\mathcal{T}$ with a virtual root
\For{each of the $I$ iterations}
    \State $h \gets$ the virtual root
    \While{$h$ cannot be further expanded} \Comment{node selection}
        \State $h \gets$ the child of $h$ with the highest UCT score \Comment{Equation~\ref{eq:uct}}
    \EndWhile
    \State $\mathcal{H}_{\mathrm{new}} \gets \mathcal{M}(\mathcal{T})$ \Comment{node expansion}
    \For{each new harness in $\mathcal{H}_{\mathrm{new}}$} \Comment{validation}
        \State add it to $\mathcal{T}$ as a child of $h$
        \State validate it on the first $r_1$ questions of $\mathcal{D}_{\mathrm{val}}$ under budget $b$
        \State compute its penalized accuracy \Comment{Equation~\ref{eq:value}}
    \EndFor
    \State update $N(\cdot)$ and $Q(\cdot)$ of all ancestors of $\mathcal{H}_{\mathrm{new}}$ \Comment{backpropagation}
    \State $h^{\dagger} \gets$ the most accurate harness not yet validated on all of $\mathcal{D}_{\mathrm{val}}$ \Comment{Equation~\ref{eq:promotion}}
    \State validate $h^{\dagger}$ on the unanswered questions of its next rung \Comment{promotion}
    \State update $\underline{\mathrm{acc}}(h^{\dagger})$ and $Q(\cdot)$ of its ancestors
\EndFor
\State \Return harnesses in $\mathcal{T}$ ranked by $\underline{\mathrm{acc}}$
\end{algorithmic}
\end{algorithm}

\section{Standard Error of Low-Fidelity Accuracy}
\label{sec:appendix_se}

We derive the bound on the standard error of the low-fidelity accuracy used in Section~\ref{sec:evolution}. Consider a harness $h$ with true accuracy $p \in [0, 1]$, and let $z_i \in \{0, 1\}$ indicate whether $h$ answers the $i$-th validation question correctly. Treating the questions as independent draws from the data distribution, each $z_i$ follows a Bernoulli distribution with mean $p$ and variance $p(1-p)$. The accuracy on the first $r$ questions is the sample mean
\begin{equation}
    \mathrm{acc}_r(h) = \frac{1}{r} \sum_{i=1}^{r} z_i,
\end{equation}
whose expectation is $p$ and whose variance is
\begin{equation}
    \mathrm{Var}\big[\mathrm{acc}_r(h)\big] = \frac{1}{r^2} \sum_{i=1}^{r} \mathrm{Var}[z_i] = \frac{p(1-p)}{r}.
\end{equation}
Since $p(1-p) = \frac{1}{4} - \big(p - \frac{1}{2}\big)^2 \leq \frac{1}{4}$, with equality at $p = \frac{1}{2}$, the standard error is bounded by
\begin{equation}
    \mathrm{SE}\big[\mathrm{acc}_r(h)\big] = \sqrt{\frac{p(1-p)}{r}} \leq \frac{1}{2\sqrt{r}}.
    \label{eq:se_bound}
\end{equation}
The bound holds for any harness regardless of its accuracy and depends only on the number of questions $r$. It equals $5.0$, $3.2$, $2.2$, and $1.6$ percentage points at $r = 100$, $250$, $500$, and $1{,}000$, respectively. The low-fidelity penalty in Equation~\ref{eq:value} is $\lambda_{\mathrm{p}}$ times this bound, so the penalized accuracy $\underline{\mathrm{acc}}(h)$ is a lower confidence bound that lies $\lambda_{\mathrm{p}}$ worst-case standard errors below the measured accuracy.

\section{Implementation Details}
\label{sec:appendix_impl}

Evolution runs on the validation set of Section~\ref{sec:valset} with a frame budget of $128$ per question. The harness proposer is Claude Code~\citep{anthropic2025claudecode} with Claude Opus 5.0, and the trace inspectors run on Claude Sonnet~\citep{anthropic2025claude}. We run for $20$ evolution iterations with $2$ children harness per iteration, producing $40$ candidate harnesses in total. For MCTS, we use an exploration weight $\lambda_{\mathrm{x}} = 0.5$. For multi-fidelity validation, we use rungs of $100$, $250$, $500$, and $1{,}000$ questions, a penalty coefficient $\lambda_{\mathrm{p}} = 1.5$, and $1$ promotion per iteration. We ensemble four harnesses in the mixture-of-harness to handle different budgets, with the low, medium, high, and xhigh tiers centered around $16$, $32$, $64$, and $128$ frames, respectively. VLM backbone decoding uses temperature $1.0$, top-$p=0.95$, and top-$k=20$.

\end{document}